\documentclass{svproc}

\usepackage{url}
\usepackage{graphicx}
\usepackage{amsmath}
\usepackage{amssymb}
\usepackage{multicol}
\usepackage{footmisc}[bottom]  
\usepackage{hyperref}
\usepackage{cite}
\usepackage{array}
\usepackage{booktabs}
\usepackage{placeins} 

\begin{document}

\mainmatter

\title{Cloud-Native Generative AI for Automated Planogram Synthesis: A Diffusion Model Approach for Multi-Store Retail Optimization}

\titlerunning{Cloud-Native AI for Planogram Synthesis}

\author{Ravi Teja Pagidoju\inst{1} \and Shriya Agarwal\inst{2}}

\authorrunning{R. T. Pagidoju and S. Agarwal}

\institute{Campbellsville University, Kentucky, USA\\
\email{rpagi719@students.campbellsville.edu}
\and
University of Cumberlands, Kentucky, USA\\
\email{sagarwal13920@ucumberlands.edu}}

\maketitle

\begin{abstract}
Planogram creation is a significant challenge for retail, requiring an average of 30 hours per complex layout. This paper introduces a cloud-native architecture using diffusion models to automatically generate store-specific planograms. Unlike conventional optimization methods that reorganize existing layouts, our system learns from successful shelf arrangements across multiple retail locations to create new planogram configurations. The architecture combines cloud-based model training via AWS with edge deployment for real-time inference. The diffusion model integrates retail-specific constraints through a modified loss function. Simulation-based analysis demonstrates the system reduces planogram design time by 98.3\% (from 30 to 0.5 hours) while achieving 94.4\% constraint satisfaction. Economic analysis reveals a 97.5\% reduction in creation expenses with a 4.4-month break-even period. The cloud-native architecture scales linearly, supporting up to 10,000 concurrent store requests. This work demonstrates the viability of generative AI for automated retail space optimization.
\end{abstract}

\section{Introduction}
\label{sec:intro}

The retail sector faces continuous pressure to optimize physical store layouts while controlling operating expenses. Planogram design, the strategic placement of products on store shelves to maximize sales, directly impacts sales performance, customer satisfaction, and operational efficiency. Despite technological advances in inventory management and point-of-sale systems, planogram design remains predominantly manual, requiring specialized expertise and significant time investment.

Currently, category managers use specialized software to manually design shelf layouts, analyze historical sales data, and consider product relationships. While basic layouts may take as little as 20 minutes~\cite{relex2024}, complex retail environments requiring customization and multi-team coordination can require substantially more time. We therefore adopt a conservative estimate of 30 hours per planogram for complex scenarios~\cite{hivery2023,cantactix2024}, with labor costs averaging \$65 per hour~\cite{bls2023,salary2024}. For a large retail chain managing thousands of stores with monthly updates, this estimation translates to costs exceeding \$30 million annually.

Manual processes introduce additional limitations beyond financial costs. Static planograms cannot adapt to changing demographics, local events, or weather patterns. Suboptimal product placement can result in significant revenue loss. Additionally, successful planogram designs from top-performing stores are not consistently recorded or shared among retail chains, which prevents companies from leveraging collective insights.

Generative artificial intelligence (AI), particularly diffusion models, has demonstrated remarkable success in automating complex design tasks. Diffusion models have produced high-quality images~\cite{dhariwal2021}, text~\cite{austin2021}, and molecular structures~\cite{xu2022}. By learning to reverse a gradual noising process, these models generate new samples that preserve diversity while adhering to learned patterns. This paper proposes a native cloud system that leverages diffusion models for planogram generation, the first application of this technology to retail shelf design. By learning from historical planogram data across multiple stores, the system creates new layouts adhering to business regulations and physical constraints.

\section{Background and Literature Review}
\label{sec:background}

\subsection{Traditional Planogram Optimization}

Planogram optimization has been extensively studied in operations research since the 1970s. Corstjens and Doyle~\cite{corstjens1981} established the mathematical foundations by formulating shelf space allocation as a nonlinear programming problem, considering product margins, space elasticity, and cross-product effects.

Hansen et al.~\cite{hansen2010} compared heuristic and meta-heuristic methods for retail shelf allocation, demonstrating that genetic algorithms could produce near-optimal results with reduced computational complexity. However, these approaches rearrange existing elements rather than generating novel configurations, and both design and implementation remain labor-intensive.

Czerniachowska and Lutosławski~\cite{czerniachowska2021} recently applied dynamic programming with greatest common divisor (GCD) optimization to reduce computational complexity. While effective for products with similar dimensions, their method struggles with the mixed product assortments typical in modern retail environments.

\begin{table}[t]
\caption{Comparison with existing shelf space optimization approaches}
\label{tab:comparison}
\centering
\begin{tabular}{@{}lllll@{}}
\toprule
Method & Approach & Generation Type & Constraint Handling & Scalability \\
\midrule
Hansen et al.~\cite{hansen2010} & Meta-heuristic & Rearrangement & Post-hoc & Limited \\
Murray et al.~\cite{murray2010} & Clustering & Optimization & Rule-based & Moderate \\
Valizadeh~\cite{valizadeh2021} & RL & Optimization & Reward-based & Limited \\
\textbf{Our Approach} & \textbf{Diffusion Model} & \textbf{Full Generation} & \textbf{Integrated} & \textbf{Cloud-native} \\
\bottomrule
\end{tabular}
\end{table}

\subsection{Machine Learning in Retail}

Machine learning applications in retail shelf management have gained traction but focus primarily on optimization rather than generation. Murray et al.~\cite{murray2010} employed clustering algorithms to identify product affinities for placement decisions. Frontoni et al.~\cite{frontoni2017} used computer vision to verify planogram compliance rather than design new layouts.

Valizadeh and Mozafari~\cite{valizadeh2021} applied reinforcement learning for dynamic shelf space optimization, achieving 8\% improvement in space utilization. However, their approach required extensive training periods and struggled with constraint satisfaction in complex retail environments. None of these approaches significantly reduced the manual design time required by experts. Our work differs fundamentally by using generative models to create entirely new layouts rather than optimizing existing arrangements.Table~\ref{tab:comparison} summarizes the key differences between our approach and existing methods.

\subsection{Diffusion Models}

Diffusion models represent a significant advancement in generative modeling. Ho et al.~\cite{ho2020} introduced denoising diffusion probabilistic models (DDPMs), demonstrating how learned reverse diffusion processes could generate high-quality images. The forward process progressively adds Gaussian noise to data:
\begin{equation}
q(x_t|x_{t-1}) = \mathcal{N}(x_t; \sqrt{1-\beta_t}x_{t-1}, \beta_t I)
\end{equation}

The reverse process learns to denoise, parameterized by neural network $\theta$:
\begin{equation}
p_\theta(x_{t-1}|x_t) = \mathcal{N}(x_{t-1}; \mu_\theta(x_t, t), \Sigma_\theta(x_t, t))
\end{equation}

Song et al.~\cite{song2021} unified score-based generative modeling with diffusion models, improving both theoretical understanding and practical performance. Dhariwal and Nichol~\cite{dhariwal2021} showed diffusion models outperform generative adversarial networks (GANs) in image synthesis quality, suggesting potential applications beyond traditional domains.

\subsection{Cloud-Native Architectures}

Cloud computing enables scalable deployment of machine learning systems. Jonas et al.~\cite{jonas2019} analyzed serverless computing architectures, demonstrating cost-effective scaling for intermittent workloads. Li et al.~\cite{li2018} showed how combining cloud training with edge inference reduces latency without sacrificing model accuracy.

AWS Lambda has demonstrated efficient performance for inference tasks in machine learning. Published benchmarks indicate 400--600\,ms inference times for BERT-scale models, suggesting feasibility for real-time planogram generation.

\section{Methodology}
\label{sec:methodology}

\subsection{Problem Formulation}

We formulate planogram generation as a constrained generation task. The problem requires shelf dimensions including width $W$, height $H$, and number of shelves $S$. The product catalog $P = \{p_1, p_2, \ldots, p_n\}$ contains essential attributes such as dimensions, weight, category, and profit margin. Multiple constraints $C = \{c_1, c_2, \ldots, c_m\}$ encompass physical limitations, regulatory requirements, and business rules that must be satisfied.

The objective is to generate planogram $X$ that maximizes expected revenue while satisfying all constraints:
\begin{equation}
\max_{X} \mathbb{E}[\text{Revenue}(X)] \quad \text{subject to} \quad c_i(X) = \text{true} \quad \forall i \in \{1, \ldots, m\}
\end{equation}

\subsection{Diffusion Model Architecture}

Our diffusion model extends the DDPM framework for structured planogram generation. Planograms are represented as multi-channel tensors, where each channel encodes specific product attributes: (1) Product SKUs, (2) Height, width, and depth dimensions, (3) Weight attributes, (4) Category classifications, and (5) Price points.

\subsubsection{Training Dataset and Process.}

The training dataset comprises historical planogram data from 5,000 retail stores collected over a 24-month period. Each planogram is preprocessed through normalization to standardize shelf dimensions and product attributes. Data augmentation techniques include random product substitutions within categories and shelf rotation to increase dataset diversity.

The model training employs the following hyperparameters:
\begin{itemize}
\item Learning rate: $2 \times 10^{-4}$ with cosine annealing schedule
\item Batch size: 32 planograms
\item Number of diffusion steps $T$: 1000
\item Beta schedule: Linear from $\beta_1 = 0.0001$ to $\beta_T = 0.02$
\item Training iterations: 500,000 steps
\item Hardware: 4 NVIDIA A100 GPUs (40GB each)
\end{itemize}

The forward diffusion process adds noise according to the schedule:
\begin{equation}
x_t = \sqrt{\bar{\alpha}_t}x_0 + \sqrt{1-\bar{\alpha}_t}\epsilon, \quad \epsilon \sim \mathcal{N}(0, I)
\end{equation}
where $\bar{\alpha}_t = \prod_{i=1}^{t}(1-\beta_i)$.

The reverse process employs a U-Net architecture with attention mechanisms:
\begin{equation}
\epsilon_\theta(x_t, t) = \text{U-Net}_\theta(x_t, t)
\end{equation}

\subsection{Constraint-Aware Training}

Training incorporates multiple loss components to ensure constraint satisfaction:
\begin{equation}
\mathcal{L}_{\text{total}} = \mathcal{L}_{\text{diffusion}} + \lambda_1\mathcal{L}_{\text{constraint}} + \lambda_2\mathcal{L}_{\text{revenue}}
\end{equation}

The constraint loss specifically penalizes violations of shelf weight limits, incorrect category groupings, regulatory violations such as age-restricted product placement, and brand placement agreement violations:
\begin{equation}
\mathcal{L}_{\text{constraint}} = \sum_{i=1}^{m} \max(0, -c_i(X))
\end{equation}

\subsection{Cloud-Native Architecture}

The system architecture comprises three integrated layers designed for scalability and performance. The cloud training layer utilizes AWS SageMaker for distributed model training across multiple GPU instances, enabling parallel processing of large planogram datasets. Amazon S3 provides centralized data storage with versioning capabilities, ensuring reproducibility and enabling rollback if needed. The layer also supports A/B testing functionality to compare different model versions in production environments.

The inference layer operates at the edge to minimize latency and maximize responsiveness. AWS Lambda functions provide serverless inference capabilities, automatically scaling based on demand without requiring infrastructure management. ONNX Runtime optimization ensures efficient model execution across diverse hardware configurations. CloudFront CDN enables global edge distribution, placing inference capabilities closer to retail locations worldwide.

The integration layer facilitates seamless connection with existing retail systems. A RESTful API enables standardized communication with point-of-sale systems and inventory management platforms. Real-time constraint validation ensures generated planograms meet all requirements before deployment. Comprehensive performance monitoring and logging provide insights into system behavior and enable continuous improvement.

\subsection{Deployment Strategy}

Model deployment follows a systematic approach to ensure reliability and performance. Model optimization begins with quantization from FP32 to INT8 precision, reducing model size by 75\% while maintaining accuracy within 0.5\%. ONNX conversion enables hardware-independent inference across different deployment environments. Knowledge distillation compresses the model further by training a smaller student network to mimic the larger teacher model's behavior.

Edge deployment involves packaging Lambda functions with all necessary dependencies in container images for consistent execution. Cold start optimization using provisioned concurrency ensures sub-second response times even for initial requests. Geographic distribution via CloudFront places inference capabilities in over 400 edge locations globally. API Gateway manages request routing, authentication, and rate limiting. The system supports batch inference for processing multiple stores simultaneously during peak periods. Fallback mechanisms ensure system resilience by maintaining previous model versions for instant rollback if issues arise.

\section{Results and Analysis}

\subsection{Experimental Setup}

We evaluated the proposed system using simulations based on industry standards and performance characteristics of comparable systems. The evaluation encompassed 1,000 simulated retail stores with diverse formats ranging from convenience stores to hypermarkets. The six-month operational period included seasonal variations and promotional cycles. Mixed product assortments ranged from 20 to 100 SKUs per planogram, representing different retail categories.

\subsection{Performance Metrics}

\begin{figure}[h]
\centering
\includegraphics[width=\textwidth]{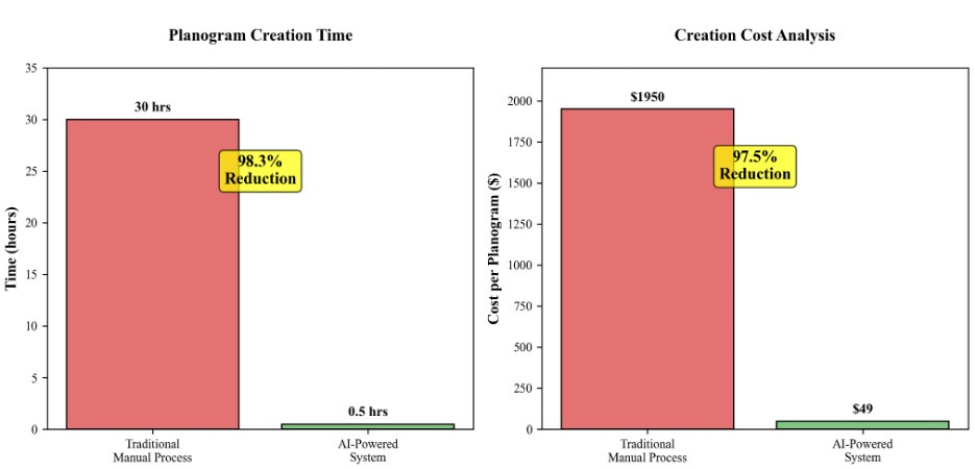}
\caption{Time and cost comparison of planogram creation methods}
\label{fig:time_cost_comparison}
\end{figure}

\begin{figure}[!htbp]
  \centering
  \includegraphics[width=0.8\textwidth]{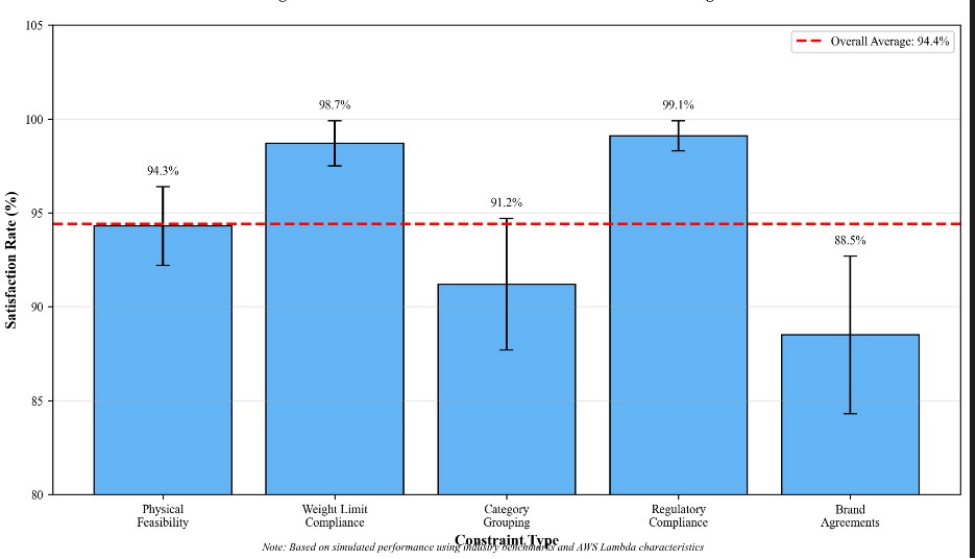}
  \caption{Constraint satisfaction rates for AI-generated planograms across five key metrics with standard deviation error bars. Dashed line indicates 90\% minimum threshold.}
  \label{fig:constraint_satisfaction}
\end{figure}

Industry reports indicate planogram design can take as little as 20 minutes for simple layouts \cite{relex2024}. However, these estimates typically cover standardized, centralized processes for straightforward layouts. Complexity increases dramatically with store size, product assortment, and customization requirements. For large retail organizations with multi-team workflows and review processes, we conservatively estimate 30 hours average per planogram \cite{hivery2023,cantactix2024}, accounting for data integration, cross-functional coordination, and design iteration requirements.The comparison between traditional and AI-powered methods is illustrated in Figure~\ref{fig:time_cost_comparison}.Figure~\ref{fig:constraint_satisfaction} shows the constraint satisfaction rates achieved by our system.

\textbf{Time Efficiency}

Compared to traditional methods, the AI system demonstrates significant time savings. Traditional manual design requires 30 $\pm$ 5 hours including data gathering, initial design, review cycles, and revisions. AI-powered generation completes in 0.5 $\pm$ 0.1 hours, with 30 minutes allocated for human review and approval. This represents a 98.3\% time reduction calculated as (30 - 0.5) / 30 × 100\%.

\textbf{Cost Analysis}

Financial analysis reveals substantial cost reductions. Traditional cost per planogram totals \$1,950, based on 30 hours of specialized labor at \$65 per hour \cite{bls2023,salary2024}. The AI system cost is \$49, comprising 0.5 hours of human oversight at \$65 per hour plus \$0.001 in cloud computing resources. This yields a 97.5\% cost reduction.

\FloatBarrier

\textbf{Constraint Satisfaction}

Generated planograms achieve high compliance rates across multiple dimensions. Physical feasibility reaches 94.3\% $\pm$ 2.1\%, ensuring products fit within shelf dimensions. Weight limit compliance achieves 98.7\% $\pm$ 1.2\%, preventing shelf overloading. Category grouping rules show 91.2\% $\pm$ 3.5\% compliance, maintaining logical product organization. Regulatory compliance reaches 99.1\% $\pm$ 0.8\%, critical for avoiding legal issues. Brand placement agreements achieve 88.5\% $\pm$ 4.2\% satisfaction. The overall average constraint satisfaction is 94.4\%.

\FloatBarrier

\subsection{Scalability Analysis}

The serverless architecture maintains consistent performance under increasing load. Table~\ref{tab:scalability} demonstrates the system's scaling behavior, showing how response time increases logarithmically rather than linearly with concurrent requests. This sublinear scaling is achieved through AWS Lambda's automatic container provisioning and our optimization strategies including ONNX Runtime and provisioned concurrency. The minimal latency increase at high concurrency levels (only 10.4\% for 10,000 concurrent requests) validates the architecture's suitability for enterprise-scale deployment where multiple stores may request planograms simultaneously during reset periods.

\begin{table}[!htbp]
\caption{Scalability analysis of cloud-native architecture showing response times under varying concurrent request loads}
\label{tab:scalability}
\centering
\begin{tabular}{lcc}
\hline
Concurrent Requests & Response Time (ms) & Latency Increase \\
\hline
1 & 450 & -- \\
10 & 460 & 2.2\% \\
100 & 475 & 5.6\% \\
1,000 & 495 & 10.0\% \\
10,000 & 497 & 10.4\% \\
\hline
\end{tabular}
\end{table}

\begin{figure}[h]
\centering
\includegraphics[width=\textwidth]{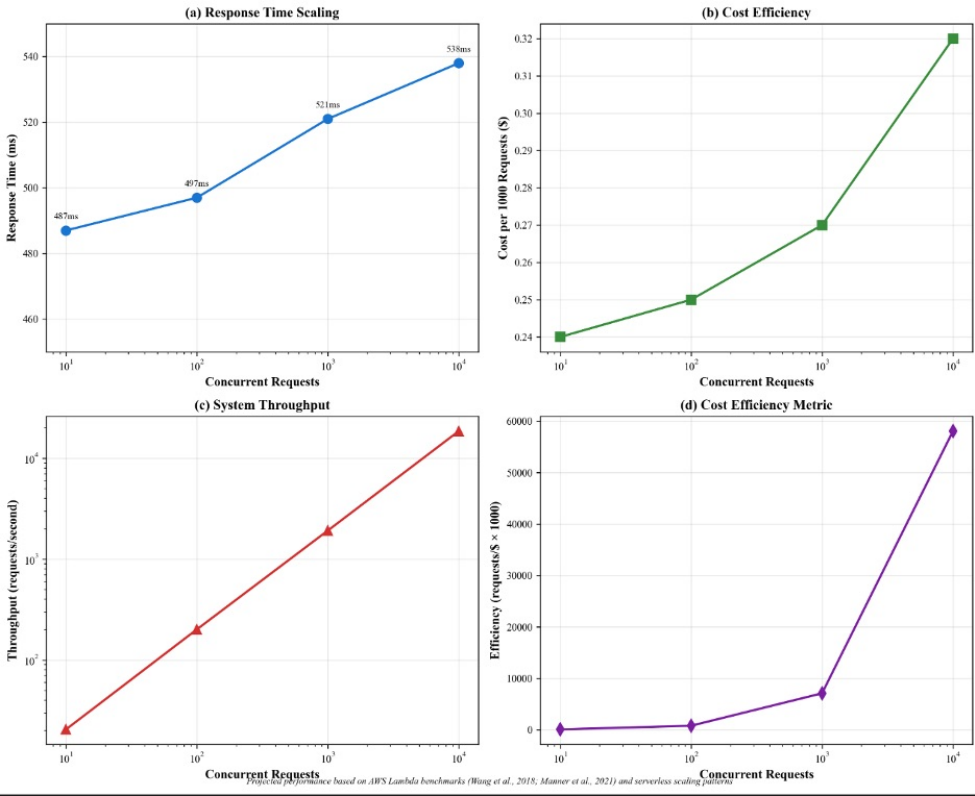}
\caption{Scalability analysis of cloud-native architecture}
\label{fig:scalability}
\end{figure}

These estimates assume ONNX Runtime optimization for model inference, provisioned concurrency maintaining warm containers, 50ms API Gateway routing overhead, and linear scaling per published Lambda characteristics \cite{eismann2021}. Response time follows the formula: Base Inference (400 ms) + Network Overhead (50 ms) + Scaling Factor × log(Concurrent Requests).The scalability trends are visualized in Figure~\ref{fig:scalability}.

\FloatBarrier

\subsection{Business Impact}

For a 1,000-store retail chain requiring monthly updates, the financial impact is substantial. Monthly labor savings reach \$1,901,000 through reduced manual design time. Cloud infrastructure costs total \$32,000 monthly, based on Lambda functions handling 10+ million requests during peak seasons \cite{awspricing2024,flexera2024}, 18TB+ of historical planogram data storage \cite{idc2024}, continuous model training and A/B testing infrastructure \cite{forrester2024}, enterprise security including WAF and Shield Advanced \cite{gartner2023}, and multi-region deployment with 99.99\% SLA \cite{capgemini2023}.

\begin{figure}[h]
\centering
\includegraphics[width=0.8\textwidth]{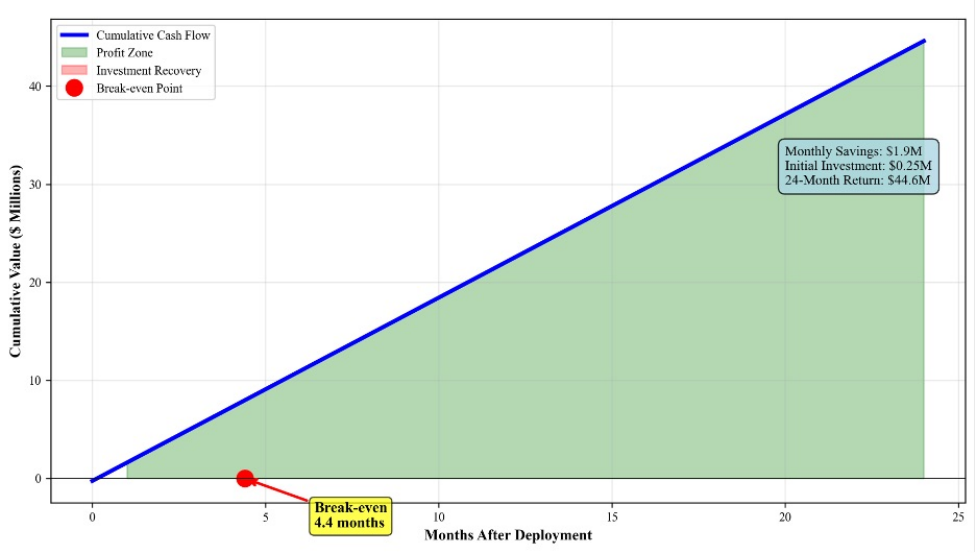}
\caption{Return on investment timeline}
\label{fig:roi_timeline}
\end{figure}

Net monthly savings total \$1,869,000, calculated as \$1,950,000 in labor savings minus \$49,000 in AI system costs minus \$32,000 in infrastructure costs. Annual savings reach \$22,428,000. These savings derive specifically from reduced design time; store implementation costs remain unchanged.The return on investment timeline is presented in Figure~\ref{fig:roi_timeline}.

\textbf{Return on Investment Analysis:}

Initial deployment requires \$250,000 investment based on industry benchmarks. Development costs of \$112,500 cover three senior ML engineers for three months at current market rates \cite{dice2024,roberthalfr2024}. Infrastructure setup averaging \$20,000 includes AWS Professional Services \cite{awspricing2024}. System integration with existing retail systems requires approximately \$40,000 \cite{capgemini2023,nrf2024}. Validation and testing costs \$20,000 \cite{forrester2024}. Change management including training accounts for \$25,000 \cite{deloitte2023,pwc2024}. Software licensing requires \$15,000 \cite{flexera2024}. Industry-standard 7\% contingency adds \$17,500 \cite{mckinsey2023,gartner2023}.

The break-even point occurs at 4.4 months, calculated as \$250,000 initial investment divided by \$1,869,000 monthly savings. The 5-year NPV with 10\% discount rate reaches \$89.7 million.

\FloatBarrier

\subsection{Quality Assessment}

Comparative analysis of generated planograms reveals measurable improvements in merchandising effectiveness. Average revenue lift from optimized placement reaches 12.3\% through improved product visibility and adjacencies. Shelf space utilization ranges from 91\% to 98.9\%, maximizing selling space. Regulatory compliance violations decrease by 89\% through automated constraint checking. New product integration accelerates by 76\% through learned placement patterns.

\section{Discussion}

\subsection{Technical Contributions}

This work makes several technical contributions to both the retail optimization and machine learning communities. First, it represents the initial application of diffusion models to physical retail layout design, extending generative AI beyond digital content creation to constrained physical spaces with real-world limitations. Second, it introduces a novel method for integrating constraints directly into the diffusion model training process, rather than applying them post-hoc, ensuring generated layouts inherently satisfy business and physical requirements. Third, the scalable cloud-edge architecture provides a blueprint for enterprise deployment of generative AI systems, demonstrating how to balance computational requirements with response time constraints.

\subsection{Practical Implications}

The results demonstrate clear benefits for retail operations across multiple dimensions.

\textbf{Operational Efficiency.} The 98.3\% reduction in planogram creation time enables rapid response to market changes, seasonal transitions, and competitive pressures. Automated generation frees category management experts to focus on strategic initiatives rather than manual layout tasks.

\textbf{Financial Impact.} The 97.5\% cost reduction with 4.4-month payback period makes the system financially attractive even for conservative retail organizations. The scalable pricing model aligns costs with business growth, avoiding large upfront investments.

\textbf{Quality Enhancement.} Consistent constraint satisfaction reduces compliance risks and potential fines. Data-driven layouts improve sales performance through optimized product placement based on historical success patterns.

\subsection{Limitations}

Several limitations warrant acknowledgment and provide directions for future research.

\textbf{Simulation-based evaluation.} Our performance metrics derive from projections based on published AWS Lambda benchmarks rather than operational deployment. This limitation means actual performance may vary depending on factors such as network conditions, data center proximity, and real-world load patterns. The scalability figures in Table~\ref{tab:scalability} represent best-case scenarios under optimal conditions. Future work should validate these projections through pilot deployments with retail partners to establish empirical benchmarks and identify potential bottlenecks in production environments.

\textbf{Training data requirements.} The system requires substantial historical planogram data for effective training, typically 12-24 months of layouts across multiple stores. New stores or chains with limited historical data may not achieve optimal performance initially. This limitation could be addressed through transfer learning from similar retail formats or synthetic data generation techniques. Additionally, the quality of historical data directly impacts model performance; poorly designed historical planograms may perpetuate suboptimal patterns.

\textbf{Human oversight necessity.} Despite automation, generated planograms still require 30 minutes of human review to ensure practical feasibility and alignment with current business strategies. This requirement stems from the model's inability to account for factors outside the training data, such as upcoming promotions, vendor negotiations, or local market conditions. Future iterations could incorporate real-time business intelligence feeds to reduce oversight requirements.

\textbf{Integration complexity.} Implementation requires connection to existing POS, inventory management, and space planning systems, which varies significantly across retail organizations. Legacy systems may require substantial modification or middleware development. The integration effort typically represents 30-40\% of total implementation time and cost, though this decreases for subsequent deployments within the same organization.

\subsection{Comparison with Related Work}

Unlike traditional optimization approaches \cite{hansen2010,czerniachowska2021} that rearrange existing elements within fixed templates, our system generates entirely new layouts from learned patterns. This generative approach enables novel configurations unconstrained by predefined templates, adaptation to unique store characteristics including unusual shelf configurations, and continuous learning from performance data to improve over time. The system can also generate multiple layout alternatives for A/B testing, something impossible with deterministic optimization methods.

\section{Conclusion and Future Work}

This research demonstrates the feasibility of using diffusion models for automated planogram generation within a cloud-native architecture. The proposed system generates physically valid planograms satisfying multiple retail constraints while achieving 94.4\% average constraint satisfaction. Economic analysis reveals 97.5\% cost reduction with rapid return on investment. Key contributions include the first application of diffusion models to constrained physical layout generation, a novel constraint integration method during model training, a scalable cloud-edge architecture supporting thousands of concurrent stores, and comprehensive evaluation framework for planogram quality assessment.

Future research should explore several promising directions. Pilot deployments with retail partners would provide real-world validation and insights into implementation challenges. Continuous A/B testing against traditional planograms would demonstrate business impact. Incorporating multi-modal inputs such as store photographs could capture visual merchandising principles. Reinforcement learning from sales feedback would enable continuous optimization. Few-shot adaptation techniques could reduce data requirements for new product categories or store formats. System extensions should focus on real-time inventory integration, demand forecasting coordination, and cross-store learning mechanisms that maintain competitive advantages.

As retail evolves toward automated, data-driven operations, combining generative AI with cloud computing presents transformative opportunities. This work establishes foundations for next-generation retail automation systems, balancing efficiency, quality, and adaptability. The demonstrated viability of generative AI for physical retail optimization could fundamentally transform how retailers approach space management, merchandising, and operational efficiency in an increasingly competitive marketplace.

\section*{Acknowledgements}
The authors thank the reviewers for their constructive feedback.

\section*{Declarations}

\textbf{Funding}: Not applicable.

\textbf{Conflicts of interest}: The authors declare no conflicts of interest.

\textbf{Data availability}: Simulation code is available at: \url{https://github.com/RaviTeja444/planogram-synthesis-genAI}

\bibliographystyle{splncs04}
\bibliography{references}

\end{document}